%% file: main.tex
\documentclass[letterpaper, 10 pt, conference]{ieeeconf}  

\IEEEoverridecommandlockouts                              

\usepackage{graphics} 
\usepackage{epsfig} 
\usepackage{mathptmx} 
\usepackage{times} 
\usepackage{amsmath} 
\usepackage{amssymb}  
\usepackage{multirow}
\usepackage{color,soul}
\usepackage{algorithm}
\usepackage{siunitx}
\usepackage{algpseudocode}

\usepackage{hyperref}
\usepackage[comma,numbers]{natbib}

\title{\LARGE \bf
Anticipatory and Adaptive Footstep Streaming \\
for Teleoperated Bipedal Robots
}

\author{
Luigi Penco$^{1}$,
Beomyeong Park$^{1}$, Stefan Fasano$^{1}$, Nehar Poddar$^{1,2}$, Stephen McCrory$^{1,2}$ \\
Nicholas Kitchel$^{1}$, Tomasz Bialek$^{1}$, Dexton Anderson$^{1}$, Duncan Calvert$^{1,2}$, Robert Griffin$^{1,2}$
\thanks{This work was funded through NASA Contract \#80NSSC20M0197 and ONR Grant N00014-19-1-2023.}
\thanks{$^{1}$The authors are with the Florida Institute for Human and Machine Cognition, 40 S Alcaniz St, Pensacola, FL 32502, United States}%
\thanks{$^{2}$The authors are with the University of West Florida, 11000 University Pkwy, Pensacola, FL 32514, United States}%
\thanks{Email : \url{lpenco@ihmc.org}}
} 

\begin{document}

\maketitle
\thispagestyle{empty}
\pagestyle{empty}

\begin{abstract}
Achieving seamless synchronization between user and robot motion in teleoperation, particularly during high-speed tasks, remains a significant challenge. In this work, we propose a novel approach for transferring stepping motions from the user to the robot in real-time. Instead of directly replicating user foot poses, we retarget user steps to robot footstep locations, allowing the robot to utilize its own dynamics for locomotion, ensuring better balance and stability.
Our method anticipates user footsteps to minimize delays between when the user initiates and completes a step and when the robot does it. The step estimates are continuously adapted to converge with the measured user references. Additionally, the system autonomously adjusts the robot’s steps to account for its surrounding terrain, overcoming challenges posed by environmental mismatches between the user's flat-ground setup and the robot's uneven terrain. Experimental results on the humanoid robot Nadia demonstrate the effectiveness of the proposed system. A video is available at \href{https://youtu.be/E_X6ol7BoyQ}{https://youtu.be/E\_X6ol7BoyQ}.
\end{abstract}

\section{Introduction}
\label{introduction}
Teleoperated humanoid robots have the potential to enable remote access and intervention in environments that are dangerous, inaccessible, or impractical for humans. Beyond practical applications, they hold promise in the entertainment industry \cite{marketus2025humanoid}. Importantly, teleoperation is also crucial for collecting high-quality physical data that powers learning-based approaches in robotics. Human operators can demonstrate tasks using the robot's body, providing real-world data that is essential for training machine learning models and robotic policies \cite{liu2023optimistic, chi2024diffusionpolicy, chi2024universal}.

One of the key challenges in robot teleoperation is achieving perfect synchronization between the user's motion and the robot's movements \cite{teleopSurvey}, particularly during fast motions. 
In our previous work \cite{bertrand2024}, we developed a system that ensures smooth robot control at high frequencies by anticipating and predicting user commands.
This system predicts and interpolates user references to match the faster update rates of the robot controller, which typically operates at a much higher frequency than the rate at which user inputs are captured.  By doing so, it prevents jittery motions that occur when user inputs are directly applied at lower frequencies ensuring seamless operation even during rapid movements. Additionally, it mitigates the challenges posed by operating the robot over standard networks, which may introduce latency and are not optimized for high-frequency control.

\begin{figure}
    \centering
    \includegraphics[width=\columnwidth]{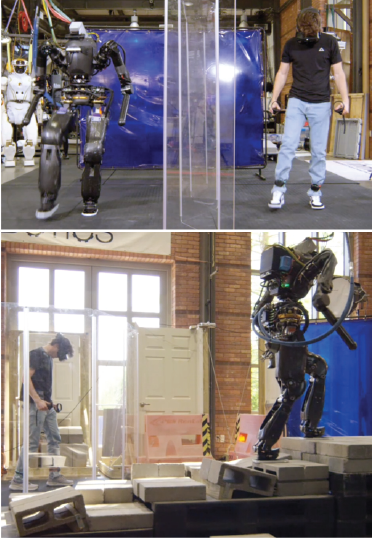}
    \caption{The humanoid robot Nadia is teleoperated by a human operator equipped with a VR headset, two VR controllers and two VR trackers. Using our footstep streaming system, which anticipates user footsteps, the robot steps in sync with the operator (top), while autonomously adapting its steps to the robot's surrounding terrain, even when the user operates on flat ground (bottom).}
    \label{fig:intro}
\end{figure}

In this work, we aim to seamlessly transfer stepping motions from the user to the robot, minimizing the delay between when the user initiates and completes a step and when the robot does it. Directly replicating user foot motions at high-speeds can easily lead to unbalanced movements due to significant differences between human and robot dynamics. For this reason, instead of directly controlling the robot's foot poses based on the user's foot movements, we propose a higher-level approach where the user only controls the footstep location of the robot. This allows the robot to utilize its own dynamics to execute steps while maintaining balance more effectively and also to adapt to the surrounding terrain. 

Research has explored different strategies to synchronize human and robot stepping in teleoperation systems. First, Ishiguro \textit{et al.} \cite{ishiguro2018} proposed a direct approach to real-time teleoperation, where the user directly controls the robot's foot pose in real time. However, due to the dynamical differences between the human operator and the robot, the user must move slowly to prevent the robot from losing balance. To ensure a safe and stable touchdown, a delay phase is introduced, allowing the robot to softly set its foot on the ground. More recent work has begun to focus on reducing delays and enhancing real-time interaction. For example, Dallard \textit{et al.} \cite{dallard2023} employ a recurrent neural network to predict walking pace and step locations. These predictive techniques show potential to minimize lag in teleoperation systems, although the the accuracy of estimation to faster human-pace motions remains to be fully validated. 

In another vein, researchers have delved into bilateral teleoperation, where human and robot dynamics are directly coupled through haptic feedback. Ramos \textit{et al.} \cite{ramos2018, ramos2019, colin2023} have developed systems that synchronize human and humanoid robot locomotion using a human-machine interface. Similarly, Ishiguro \textit{et al.} \cite{ishiguro2020} introduced a full-body exoskeleton cockpit for bilateral teleoperation, aiming to seamlessly align the movements of the operator with those of the robot. These methods are promising for enhancing teleoperation by dynamically aligning human and robot motions. However, they face challenges such as sensitivity to network delays, which can compromise real-time performance, the requirement for large, complex user-side structures that can be cumbersome and maintenance-intensive. Furthermore, the operator's responsibility for maintaining the robot's balance can potentially overload them mentally, especially in scenarios where the robot's environment is dynamic.

Another critical consideration is that, in many real-world scenarios, the environment at the robot's location cannot be mirrored at the user's side, creating significant challenges for teleoperation. This disparity makes it difficult for the user to accurately specify where the robot's foot should land. For example, while the user may operate from a flat ground setting, the robot might encounter stairs, uneven terrain, or obstacles at varying heights that the user cannot physically climb or descend. This mismatch in environments adds complexity to teleoperating bipedal robots and necessitates systems capable of interpreting high-level user commands while autonomously adapting to local terrain conditions.

Planning over rough terrain has predominantly been developed for autonomous robot navigation, where users specify high-level goals, such as a final stance location, without requiring continuous motion input \cite{griffin2019, mccrory2023, calvert2022}. Learning-based methods, such as \cite{long2024}, utilize onboard elevation maps to optimize locomotion, enabling humanoid robots to traverse stairs, gaps, and high platforms. Also Zhuang \textit{et al.} \cite{zhuang2025} propose a unified vision-based learning system for humanoid robots, allowing the robot to traverse terrains that require jumping on platforms and leaping over gaps. However, these approaches do not address how robots adapt their motion to the terrain based on direct user input. This limitation is critical in real-world dynamic scenarios where human decision-making is essential. Humans can identify hazards that AI might overlook --for instance, recognizing that an apparently stable region made of moldy wood is unsafe to step on. Similarly, in highly complex navigation tasks, humans can plan paths that simplify traversal and reduce risks more effectively than autonomous systems.

In this work, we present a solution that simultaneously addresses the two key challenges of reducing delay of operation between the user and the humanoid during stepping and adapting the steps --retargeted from the user to the robot-- to the terrain surrounding the robot.

\section{Footstep Retargeting from User to Robot}
\label{sec:ret}

 The system continuously monitors the ankle trackers to identify when a step is initiated or completed. The instantaneous direction of each step is determined directly from the latest tracker data.

To estimate stride length, we blend the user’s horizontal displacement and average velocity, ensuring that the estimate converges to the actual measured displacement at foot landing. Simultaneously, we estimate yaw turns by combining the measured rotation with an average yaw rate, applying a similar blending approach.

Finally, these user-centric translations and rotations are transformed into robot foot placements within the stance frame, subject to constraints that guarantee kinematically feasible steps. The resulting footsteps are then sent to the robot controller, as described in Section \ref{sec:control}.

\subsection{Step Detection}
We continuously monitor each foot's \emph{horizontal displacement} and \emph{vertical lift} relative to its position at the start of a step. Once the displacement and lift exceed predefined thresholds,
\begin{equation*}
  \|\mathbf{p}_{\mathrm{current}}^{XY} - \mathbf{p}_{\mathrm{initial}}^{XY}\| 
  \;\geq\; \delta_{\mathrm{step}}, 
  \quad
  (p_{\mathrm{current}}^{z} - p_{\mathrm{initial}}^{z}) 
  \;\geq\; \delta_{\mathrm{lift}},
\end{equation*}
we declare a new step is beginning and store the initial transforms for subsequent estimates. If, later, movement falls below a stability threshold for a number of iterations, we conclude the step is finished.

\subsection{Direction Update via Instantaneous Measurement}

At each iteration, we compute the \emph{instantaneous foot direction} directly from the latest tracker reading. Let $\mathbf{p}_{\mathrm{tracker, current}}^{XY}$  and $\mathbf{p}_{\mathrm{tracker, initial}}^{XY}$ be the current and initial 2D positions (horizontal plane) of the user’s foot tracker, respectively. We form the \emph{direction} as:
\begin{equation*}
  \mathbf{d}_{\mathrm{instant}}^{XY}
  \;=\;
  \frac{\mathbf{p}_{\mathrm{tracker, current}}^{XY}
    \;-\;
    \mathbf{p}_{\mathrm{tracker, initial}}^{XY}}
  {\Bigl\|\mathbf{p}_{\mathrm{tracker, current}}^{XY}
    \;-\;
    \mathbf{p}_{\mathrm{tracker, initial}}^{XY}\Bigr\|}.
\end{equation*}


\subsection{Step Length Estimate}
\label{subsec:strideScaling}

We estimate the user’s \emph{stride length} by blending information from horizontal displacement and velocity with vertical motion indicators (i.e., foot height and vertical velocity). Our approach proceeds as follows:

\subsubsection{Horizontal-Based Stride Estimate}

Let \(d_{M}\) denote the current horizontal displacement of the user’s swing foot from its initial position, and define \(\overline{v}_{\mathrm{H}}\) as the average horizontal velocity of the foot over the ongoing step cycle.
The remaining time in the robot's current step, denoted as \(\Delta t_{step}\), is calculated as the difference between the total duration of the step\footnote{The robot's step duration is configured to match the fastest robust stepping speed achievable by the robot. This accounts for the fact that humans are significantly faster at stepping than robots. Even at moderate speeds, human steps are difficult for robots to replicate due to mechanical limitations and the stability requirements of robotic systems.} and the elapsed time since the step began.
This is used to compute a raw stride length
\begin{equation*}
  L_{\mathrm{raw}}
  \;=\;
  d_{M}
  \;+\;
  \overline{v}_{\mathrm{H}} \,\Delta t_{step},
  \label{eq:rawStrideHoriz}
\end{equation*}
reflecting how much further the foot is expected to travel if it continues at its average horizontal speed for the remainder of the step. If \(\overline{v}_{\mathrm{H}}\) is large, the user likely continues forward motion, inflating the stride estimate; if small, we rely more on \(d_{M}\).

\subsubsection{Landing Factor from Vertical Position}

During stepping, to establish when the stride estimate should converge to the raw measured \(d_{M}\), we continuously check whether the foot is descending by verifying if its vertical velocity becomes negative \(\dot{z} < 0\). If so, we define a \emph{landing factor}, \(\lambda\in[0,1]\), based on the foot’s current height \(z\) relative to a maximum observed height \(z_{\max}\). Specifically:
\begin{equation*}
  \lambda 
  \;=\;
  \max\!\Bigl(0,\;\min\!\bigl(1,\;z/z_{\max}\bigr)\Bigr).
  \label{eq:landingFactorVertical}
\end{equation*}
If \(z \approx z_{\max}\), then \(\lambda\approx1\) and no vertical-based reduction is applied (the foot is still high). If \(z\) is near 0, then \(\lambda\approx0\), implying the foot is near the ground and the step is nearly complete. When \(\dot{z}\ge0\), we skip this reduction by setting \(\lambda=1\).

\subsubsection{Combining Horizontal and Vertical Cues}

We linearly blend \(L_{\mathrm{raw}}\) from \eqref{eq:rawStrideHoriz} with \(d_{M}\), the measured distance, using the factor \(\lambda\) from \eqref{eq:landingFactorVertical}:
\begin{equation*}
  L_{\mathrm{blended}}
  \;=\;
  \lambda \; L_{\mathrm{raw}}
  \;+\;
  (\,1 - \lambda\,)\,d_{M}.
\end{equation*}
Hence, when the foot is high or not descending (\(\lambda\approx1\)), the stride length remains near \(L_{\mathrm{raw}}\). If the foot is descending (\(\dot{z}<0\)) and \(z \ll z_{\max}\), then \(\lambda \to 0\), pulling the stride back toward the already traveled distance \(d_{M}\). 

\subsubsection{Clamping and P-Control} 

After blending, we clamp \(L_{\mathrm{blended}}\) within \([0,\,L_{\max}]\) to prevent negative or excessively large steps. We then apply a proportional control law to update the final stride estimate 
\[
  L_{\mathrm{est}} 
  \;\leftarrow\;
  L_{\mathrm{est}}
  \;+\;
  k_{P,\mathrm{stride}}
  \bigl(\,L_{\mathrm{blended}} - L_{\mathrm{est}}\bigr),
\]
ensuring a smooth convergence over time, and clamp the result again as needed.
We only use a P-controller and not a PD-controller (more common in robotics) because adjusting the stride length (instead of controlling a motor or joint) typically has lower overall dynamics. Small estimate changes can be handled effectively with a proportional correction, making a derivative term unnecessary for footstep stride updates.

This combination of average horizontal velocity (for predicting continued motion) and vertical-position-based landing factor (for reducing stride if the foot is near touchdown) provides a stable yet adaptable stride estimate in real-time operation.

\subsection{Yaw (Turning) Estimate}
We similarly track a running average of the foot’s yaw-rate (\(\dot{\psi}\)). The \emph{measured} yaw change \(\Delta\psi_{\mathrm{measured}}\) is the difference between the current and initial foot yaw for this step. A raw yaw estimate is
\begin{equation*}
  \psi_{\mathrm{raw}} 
  \;=\;
  \Delta\psi_{\mathrm{measured}}
  \;+\;
  \overline{\dot{\psi}}\;\Delta t,
\end{equation*}
which is again blended with the measured value if the foot is descending (same \(\lambda\) logic as above). A P-control step and clamping to \(\pm\psi_{\max}\) yield the final \(\psi_{\mathrm{est}}\).

\subsection{Final Footstep Composition}
The final foot placement combines the direction \(\mathbf{d}_{\mathrm{instant}}^{XY}\), stride length \(L_{\mathrm{est}}\), and yaw \(\psi_{\mathrm{est}}\). Specifically, the foot is offset in the XY-plane by  $L_{\mathrm{est}}\,\mathbf{d}_{\mathrm{instant}}^{XY}$,
and rotated about the vertical axis by \(\psi_{\mathrm{est}}\) from its initial stance reference.

After determining the intended translation and yaw of the user's tracker, we map these tracker-centric measurements onto the robot frames to produce a valid footstep. 
By synthesizing the local measured foot translation and rotation of the user into the robot stance foot frame, we obtain a coherent real-time footstep command that couples the user stance to that of the robot. Furthermore, we clamp both the lateral distance and the relative yaw rotation with respect to the stance foot, so that potentially unsafe maneuvers (e.g., cross-stepping or internal foot rotations) are disallowed under teleoperation, even though the robot may autonomously use such configurations for balance recovery if needed. Such restrictions keep teleoperated commands intuitive and within physically sensible limits for the intended user step.

\section{Footstep Adjustment}
\subsection{Height Map Generation}
\label{sec:map}
A GPU-based pipeline was designed and developed to construct elevation maps of the terrain in real-time.  The elevation map is defined as a scalar-valued function \( ^F{h^t} : \mathbb{R}^2 \rightarrow \mathbb{R} \), which assigns a height value to every 2D location on the terrain at time \( t \), referenced in a global frame \( F \).

The generation pipeline consists of two main stages: local extraction and global registration.  Utilizing an NVIDIA RTX 4060 GPU, the algorithm achieves an average computation time of 5 milliseconds for data acquired at a 30 Hz capture rate.
\subsubsection{Local Height Map Extraction}
In the local extraction stage, an instantaneous elevation map is computed in a sensor-aligned local frame \( \mathcal{L} \), which shares the world frame's $\mathcal{W}$ $XY$-plane but is yaw-aligned with the sensor pose to optimize memory access patterns and avoid synchronization overhead on the GPU. Each cell of the local grid is processed in parallel on the GPU, where a thread computes the height by projecting the cell's location into the camera frame using the camera's intrinsic parameters, and then averaging the heights of valid depth points within a defined neighborhood. This is formalized as:
\[
h_l(p) = \frac{1}{N} \sum_{i=0}^{N} \pi^{-1}(\pi(p) + C_p)
\]
where \( p \) denotes the 2D location of the cell in the local heightmap grid, and \( C_p \) is an offset vector that indexes into a neighborhood around \( p \) in the image plane. The function \( \pi \) represents the camera's perspective projection, mapping 3D points to 2D image coordinates, while \( \pi^{-1} \) is its inverse, mapping image coordinates (with depth) back to 3D space. 
The perspective projection function $\pi : \mathbb{R}^3 \rightarrow \mathbb{R}^2$ is more formally defined as
\begin{equation}
\pi(\mathbf{x}) = 
\begin{bmatrix}
u \\
v \\
1
\end{bmatrix}
=
\frac{1}{z}
\begin{bmatrix}
f_x & 0 & 0 & c_x \\
0 & f_y & 0 & c_y \\
0 & 0 & 0 & 1
\end{bmatrix}
\begin{bmatrix}
x \\
y \\
z \\
1
\end{bmatrix}
\label{eq:perspective_projection}
\end{equation}
where $f_x$ and $f_y$ are the focal lengths, and $c_x$ and $c_y$ are the image center offsets of the depth camera.

To prevent erroneous height readings caused by the robot's own body, the collision meshes of the robot's arms and legs are also used as a filter for the height map. Any height measurements at locations where the robot's body parts would be detected are excluded from registration in the global height map.

\begin{figure}
    \centering
    \includegraphics[width=\columnwidth]{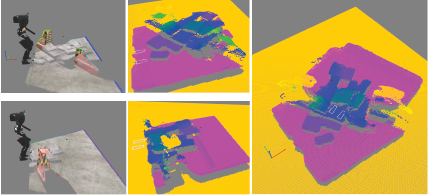}
    \caption{Local elevation map extraction and quasi-global registration. The local elevation maps (center top and bottom) are combined to get the global elevation map (right).}
    \label{fig:heightmap}
\end{figure}

\subsubsection{Quasi-Global Map Registration}
Local height maps are computed at high rates (125--200 Hz), but are limited by occlusions and the sensor's field of view. To address this, a quasi-global registration step fuses the local map into a persistent global height map in world frame ($h_\mathcal{W}(p)$), using a complementary filter (Figure \ref{fig:heightmap}):
\[
h_\mathcal{W}(p) = E(\alpha h_\mathcal{W}^{(t-1)}(p) + (1-\alpha) ^\mathcal{L}h^{(t)}(^\mathcal{L}T_\mathcal{W}p))
\]
where \( \alpha \) is a filter parameter for selecting the contribution of the incoming measurement to the final estimate and \( ^\mathcal{L}T_{\mathcal{W}} \) transforms local coordinates to the world frame. This fusion is performed on the GPU for efficiency.

Once we have the global world-frame map, we perform filters on the global world-frame map to reduce overall noise. First, we perform an exponentially moving average filter on each cell in the global world-frame map. Then we perform a filter to remove points that are too far from the mean of the surrounding cells. If a point is marked for removal, it gets replaced by the mean of the surrounding cells. The resulting global map is stored persistently in memory and updated using a dedicated GPU kernel.

\subsection{GPU-Accelerated Terrain Adaptation}
We optimize footstep placement on uneven terrain using a GPU-accelerated approach. The system integrates a grid search over potential footstep poses, evaluates terrain planarity, and ensures stability through a composite cost function.

\subsubsection{Decision Variables and Search Space Configuration}
The optimization process searches for an optimal footstep pose $\mathbf{q} = \begin{bmatrix} x & y & \theta \end{bmatrix}^\top$ within a predefined search space:
\begin{equation*}
\begin{aligned}
    x, y &\in [p^* - 1.5f_x, p^* + 1.5f_x], \\
    \theta &\in [\theta^* - \frac{\pi}{4}, \theta^*  + \frac{\pi}{4}],
\end{aligned}
\end{equation*}
where $f_x$ is the length of the robot foot, and $p^*$ and $\theta^*$ represent the target planar position and yaw, respectively. The spatial resolution of the search directly depends on the height map's resolution (2cm), as it determines how finely terrain heights are sampled and evaluated. For angular resolution, the yaw orientation of the foot is sampled at increments of 5 degrees. The search radius is defined as 1.5 times the foot length, ensuring that the optimization considers a sufficiently large area around the target position. Furthermore, the maximum yaw deviation allowed for candidate poses is limited to ±45 degrees.
This search space structure maps directly to CUDA thread blocks for efficient parallel computation.

\subsubsection{Cost Function}
The objective of optimization is to minimize a composite cost function that balances proximity to the target pose with the stability of the terrain. The cost function is defined as:
\begin{equation*}
\begin{aligned}
    \mathcal{J}(\mathbf{q}) = &\ w_{\text{position}}||\mathbf{p}-\mathbf{p}^*||_1 + 
    w_{\text{yaw}}|\theta-\theta^*| \\
    &+ w_{\text{planarity}}\Phi(\mathbf{q}) + 
    w_{\text{height}}|z-z^*|,
\end{aligned}
\end{equation*}
where the first term penalizes deviations from the target position $\mathbf{p}^*$; the second term penalizes deviations from the target orientation $\theta^*$; the third term, $\Phi(\mathbf{q})$, evaluates the planarity of the terrain under the foot; the fourth term penalizes deviations in height from the terrain elevation at the target position $z^*$.
Weighting factors are chosen to prioritize stability while maintaining reasonable proximity to the target pose ($w_{\text{position}} = 10$, $w_{\text{yaw}} = 30$, $w_{\text{planarity}} = 100$, $w_{\text{height}} = 1$).

\subsubsection{Planarity Assessment}

The planarity cost $\Phi(\mathbf{q})$ evaluates the stability of a candidate footstep pose by analyzing the terrain underfoot. This process involves sampling terrain heights at key points, checking for surface continuity, and fitting a plane to evaluate deviations and slopes. The following points describe each step in detail.

\subsubsection*{Sampling Strategy}
To assess the stability of the terrain, five key points are sampled on the surface of the foot: one in each corner of the foot and one in its center. The positions of these points are determined based on the foot's dimensions and orientation:
\begin{equation*}
    \mathcal{V} = 
    \left\{
        \mathbf{p}_c + R(\theta)\begin{bmatrix}\pm f_x/2 \\ \pm f_y/2\end{bmatrix}, 
        \mathbf{p}_c
    \right\},
\end{equation*}
where $\mathbf{p}_c$ is the center position of the foot, $f_x$ and $f_y$ are its length and width, respectively, and $R(\theta)$ is the rotation matrix for yaw orientation $\theta$. The heights at these points are extracted from a heightmap representing the terrain.

\subsubsection*{Surface Continuity Check}
Before proceeding with a further analysis, a surface continuity check is performed to detect abrupt elevation changes. The average height of the front corners (points 1 and 4) and back corners (points 2 and 3) is computed as:
\begin{equation*}
    h_{\text{front}} = \frac{h_1 + h_4}{2}, \quad 
    h_{\text{back}} = \frac{h_2 + h_3}{2}.
\end{equation*}
The center height $h_5$ is compared to the mean of these averages:
\begin{equation*}
    h_{\text{mean}} = \frac{h_{\text{front}} + h_{\text{back}}}{2}.
\end{equation*}
If the absolute difference between $h_5$ and $h_{\text{mean}}$ exceeds a predefined threshold the surface is deemed discontinuous and a penalty proportional to height differences between each corner and the center point is applied:
\begin{equation*}
\begin{aligned}
    P_{\text{discontinuity}} = w_{discontinuity} \cdot 
    ( |h_1 - h_5| + |h_2 - h_5| + \\
    |h_3 - h_5| + |h_4 - h_5| ).
\end{aligned}
\end{equation*}

\subsubsection*{Plane Fitting and Steppability Metrics}
If the terrain underfoot is determined to be continuous, a plane is fitted to the sampled points to evaluate the terrain's suitability for foot placement. The plane fitting is performed using a least squares method, which minimizes the sum of squared errors between the sampled point heights and the fitted plane. This process calculates the coefficients of the plane directly from the sampled positions and heights.

Once the plane is fitted, several metrics are computed to assess the terrain's planarity and steppability. The first metric is the mean height deviation $\bar{\delta}_{height}$, which measures the average distance between each sampled point height and the fitted plane. This provides an overall indication of how well the terrain conforms to a flat surface. The second metric is the maximum height variance $\sigma_{height, \text{max}}$, which captures the largest deviation among all sampled points from the fitted plane. This metric highlights any significant irregularities in the terrain. The third metric is the maximum slope of the fitted plane $\phi_{\text{max}}$, which quantifies its steepest inclination. The slope is calculated as:
\begin{equation*}
    \phi = \arctan{\sqrt{\alpha^2 + \beta^2}},
\end{equation*}
where $\alpha$ and $\beta$ are coefficients representing the tilt of the plane in the x and y directions, respectively.

If any of these metrics exceed predefined thresholds, penalties are applied to discourage selecting unstable or unsuitable foot placements. A height variance penalty $P_{\sigma}$ is added if the maximum variance exceeds 5 cm, while a slope penalty $P_{\phi}$ is applied if the maximum slope exceeds 50 degrees.

The total planarity cost is expressed as:
\begin{equation*}
    C_{\text{planarity}} = \bar{\delta}_{height} + P_{\phi} + P_{\sigma},
\end{equation*}
or 
\begin{equation*}
    C_{\text{planarity}} = P_{discontinuity}
\end{equation*}
based on whether the terrain is continuous or not.

\subsubsection{GPU-Accelerated Optimization Pipeline}

The optimization process is implemented as a three-stage pipeline on an NVIDIA RTX 4060 GPU to achieve real-time performance. First, each CUDA thread evaluates the cost function \(\mathcal{J}(\mathbf{q})\) for a single candidate footstep pose within the search space. Memory transfers between the host (CPU) and device (GPU) are overlapped with computation to ensure efficient data handling.
Next, a block-wise reduction is performed within each thread block using shared memory to identify local minima of the cost function. This step reduces computational overhead by narrowing down candidate solutions.
Finally, a global minimum search aggregates the local minima from all blocks and determines the optimal footstep pose.

This implementation evaluates over \(10^5\) candidate poses in under 4 milliseconds, achieving real-time performance at approximately 285 Hz. This high frequency aligns well with virtual reality input rates (60-120 Hz), ensuring responsive footstep optimization.

\section{Executing footsteps}
\label{sec:control}
Given a real-time stream of updated footsteps, the robot performs whole-body motion through a layered control architecture that prioritizes stepping stability. The controller dynamically adjusts joint trajectories and contact forces to achieve commanded foot placements while maintaining balance. 

\begin{figure}
    \centering
    \includegraphics[width=\columnwidth]{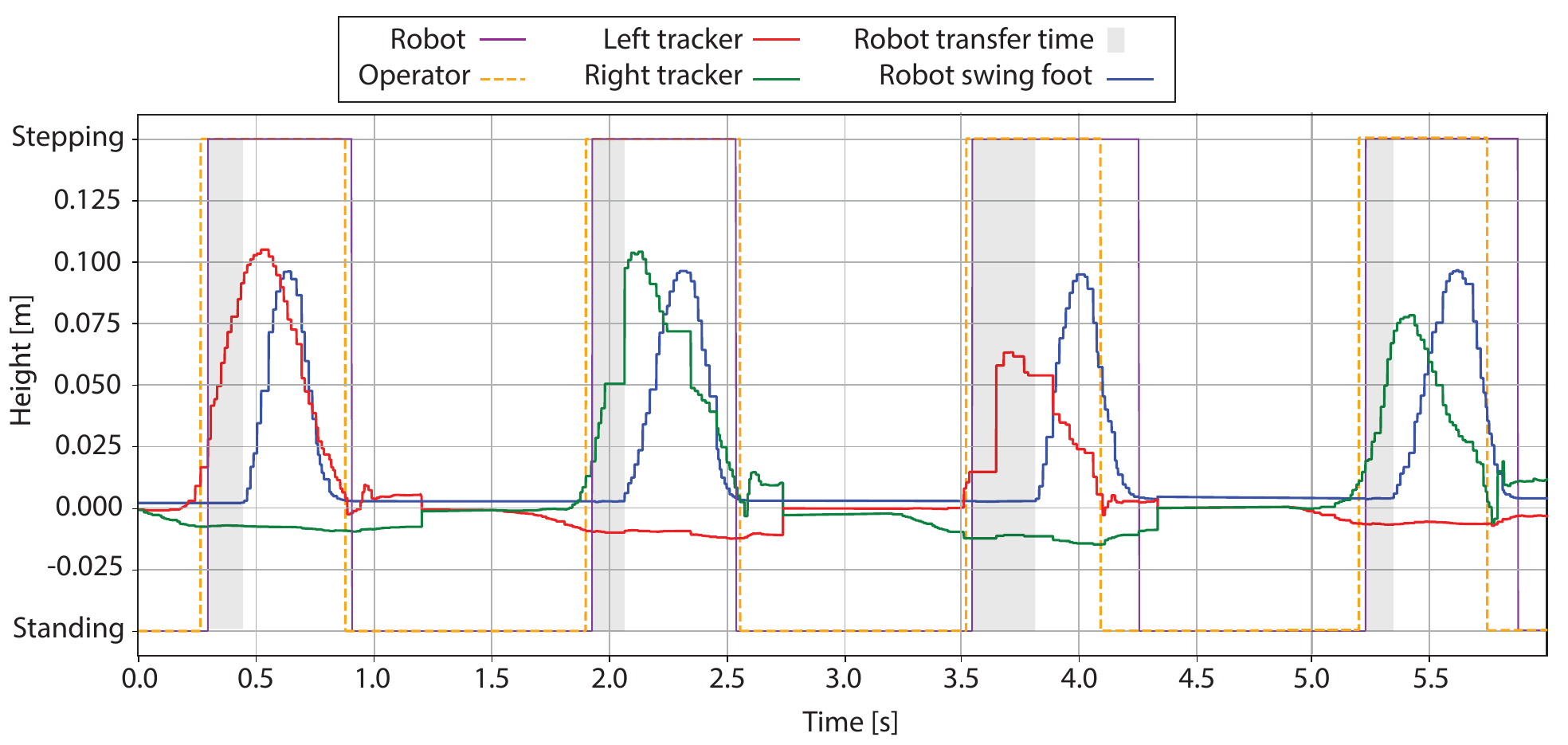}
    \caption{Synchronization of user and robot stepping during real-time footstep streaming. Each user step triggers an immediate footstep command to the robot, which initiates its own step after a brief transfer phase.}
    \label{fig:timeStep}
\end{figure}

\begin{figure*}
    \centering
    \includegraphics[width=\linewidth]{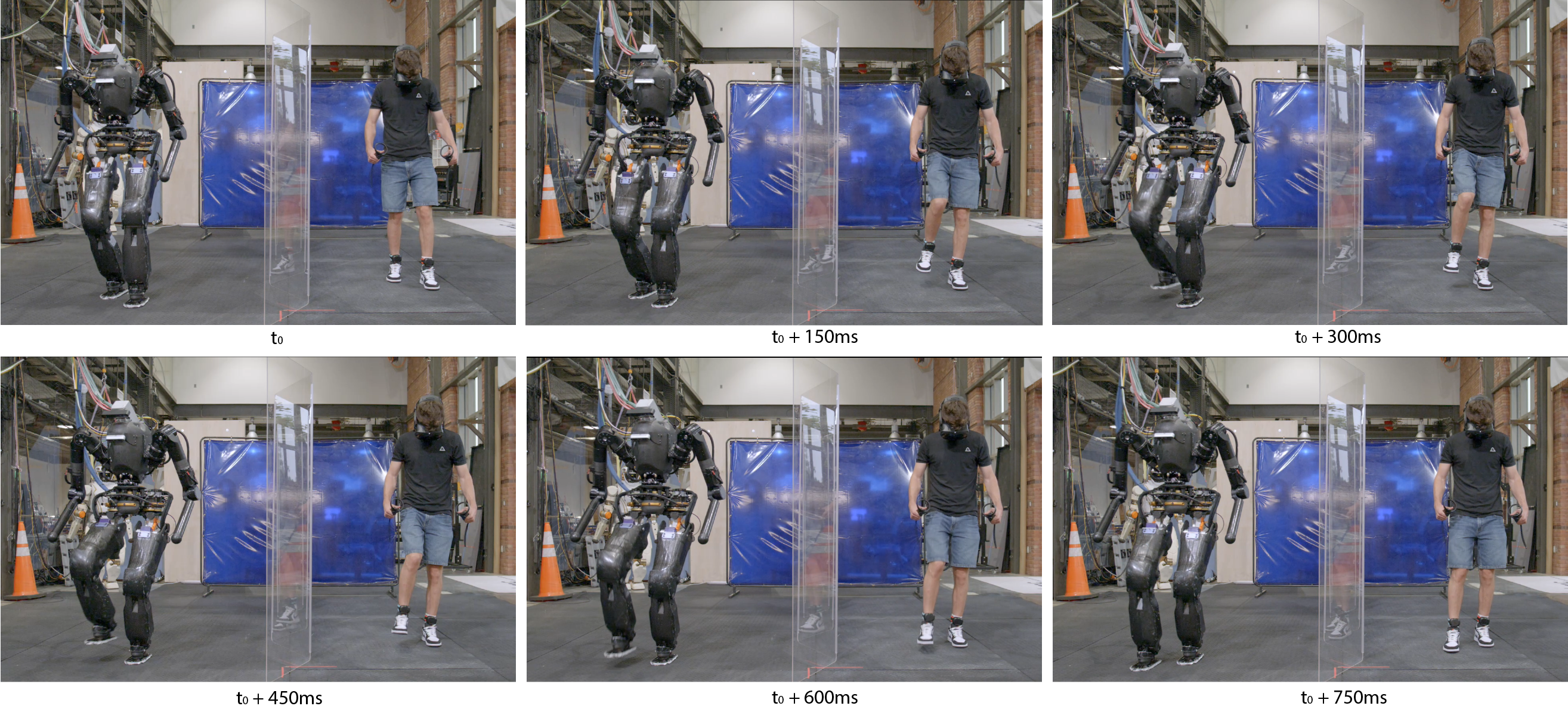}
    \caption{Screenshots of the user teleoperating the robot to step on flat ground. The images show the user’s steps being mirrored by the robot, demonstrating responsive footstep execution during teleoperated walking on level terrain.}
    \label{fig:screenshots}
\end{figure*}

\subsection{Stepping controller}
The walking controller employs a momentum-based whole-body quadratic programming (QP) framework that simultaneously tracks desired momentum changes, and satisfies kinematic objectives and contact constraints \cite{koolen2016}. This architecture integrates three operational phases through a state machine: standing, transfer, and swing. During standing, full foot contact maintains static equilibrium through constrained optimization. The transfer phase initiates weight shifting by generating instantaneous capture point (ICP) trajectories \cite{pratt2006capturepoint} while relaxing stance foot constraints to permit natural heel lift. Subsequent swing phase execution employs quintic spline trajectories for precise foot placement, guided by real-time updates from the footstep stream.

Balance management operates through ICP-driven trajectory generation  \cite{Pratt2019_bookchapter}. The QP solver outputs joint accelerations and contact wrenches, which are converted to actuator torques via inverse dynamics. This enables continuous compliance adjustments during single support phases while preserving the viability of user-commanded footsteps.

Upper body coordination maintains default torso stationarity during locomotion, with optional user-specified motions superimposed through independent end-effector pose tracking. For a comprehensive understanding of the balance dynamics and control strategies employed, additional details can be found in \cite{koolen2016}.

\input{SwingTrajectoryCollisionAvoidance}

\section{Results}
We conducted experiments with Nadia, our humanoid robot provided with 21 degrees of freedom and partially powered by hydraulics. To operate the robot, users were equipped with a Valve Focus 3 VR headset and controllers, together with a Vive Ultimate tracker attached to each ankle. 

To evaluate the system’s capability for real-time footstep streaming--where footstep commands are issued to the robot as soon as the user steps--we performed trials at a typical human walking pace on flat ground.

Figures \ref{fig:timeStep} and \ref{fig:screenshots} illustrate the synchronization between the user's steps and the robot's movements. Each time the user initiates a step, a corresponding footstep command is immediately sent to the robot, prompting it to begin its own step. To keep pace with normal human walking speeds, the robot was configured to step as quickly as its actuators would allow.

However, unlike a human, the robot does not lift its foot into the swing phase immediately after starting to step. Instead, the employed controller (Section \ref{sec:control}) requires the robot to first enter a transfer state, which we set to the fastest achievable duration of 0.15 seconds. In practice, the robot does not always achieve this minimum; depending on the ICP error and the stance configuration, the transfer phase may extend up to 0.25–1 seconds. These timing variations are evident in Figure \ref{fig:timeStep}, highlighting the robot’s transfer time in different steps.

As the user initiates a step, an initial estimate for the robot’s corresponding footstep is generated using the method described in Section \ref{sec:ret}. This estimate is not static; instead, it is continuously refined as the user progresses through their step, ensuring that the robot’s movement remains closely aligned with the user’s intent.

\begin{figure}
    \centering
    \includegraphics[width=\columnwidth]{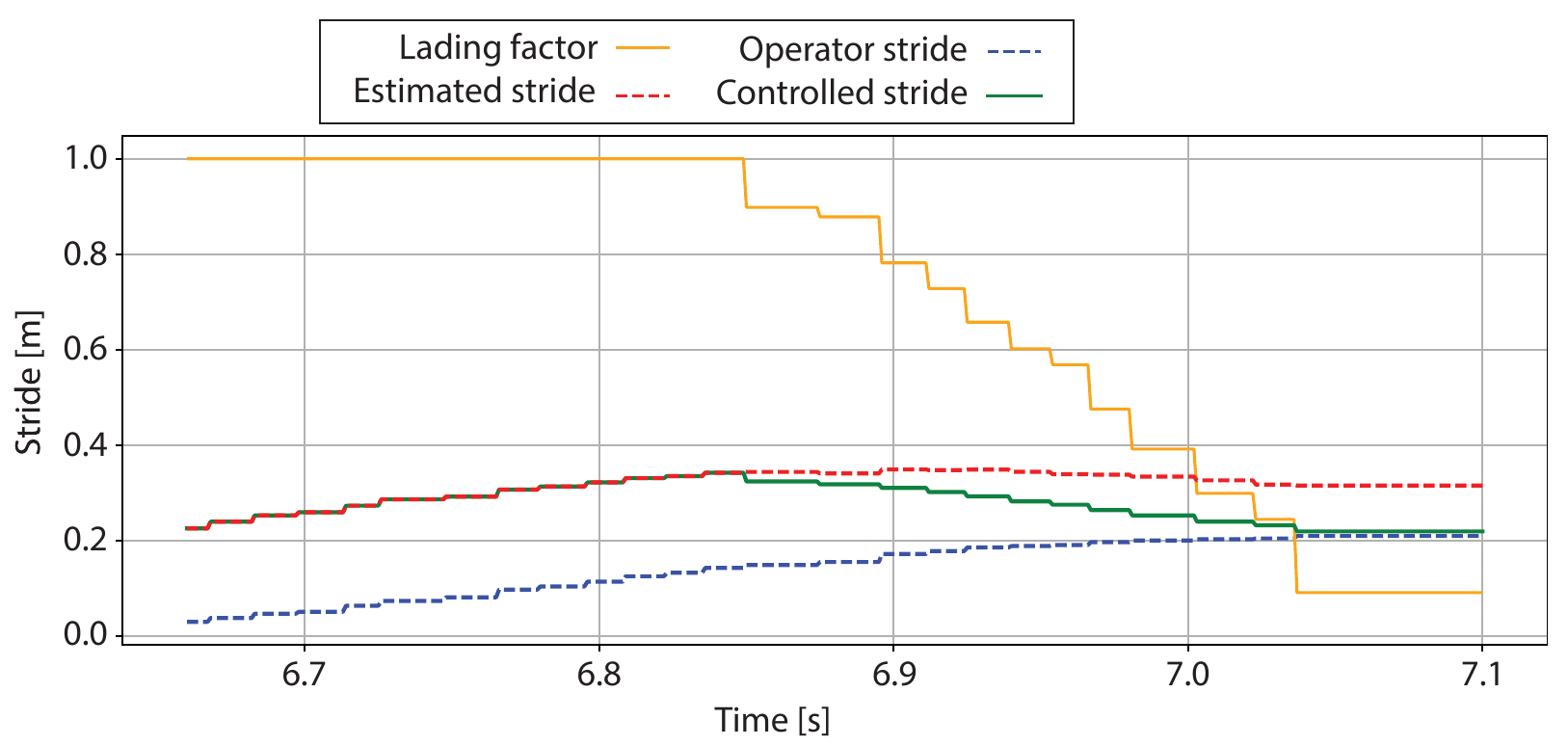}
    \caption{Evolution of the robot’s controlled footstep estimate throughout a user’s stride. The initial estimate is continuously refined in real time as the user’s foot moves, converging to the true landing position by the end of the step. This dynamic update mechanism enables precise and responsive teleoperated robot stepping.}
    \label{fig:stride}
\end{figure}

Figure \ref{fig:stride} illustrates how the controlled footstep--the command ultimately sent to the robot controller--is dynamically updated in real time. The raw initial estimate typically provides a reasonable approximation of the user’s final stride location. However, as the user’s foot moves toward its landing position, the estimate is further adjusted to more accurately reflect the actual placement of the user’s foot.

As the landing factor approaches zero, the controlled footstep converges to the user’s true landing position. The evolution of the footstep estimate over time, as depicted in Figure \ref{fig:stride}, demonstrates the effectiveness of this continuous update mechanism in achieving accurate and responsive teleoperated robot stepping.

\begin{figure*}
    \centering
    \includegraphics[width=\linewidth]{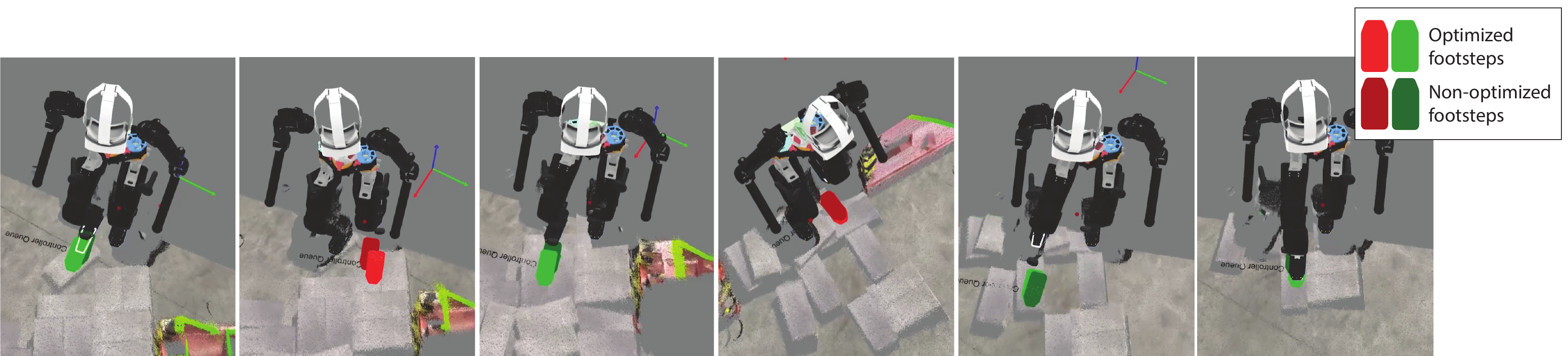}
    \caption{Visualization of footstep adjustment on rough terrain. The robot dynamically corrects--while stepping-- user-indicated steps that fall on unstable regions, projecting them onto safe, planar areas of the cinder block hill for reliable traversal.}
    \label{fig:roughTerrainUI}
\end{figure*}

We have also evaluated the robot’s ability to adjust its footsteps based on the perceived terrain. We teleoperated the robot to climb a cinder block hill (Figure  \ref{fig:intro}, optimizing user retargeted steps (Section \ref{sec:ret}), using a height map reconstructed from depth sensing onboard (Section \ref{sec:map}). To ensure stable footing on the uneven surface, we reduced the robot’s walking speed. When the intended step of the user was landed on the edge of a cinder block or in an non-steppable area, the system's optimization automatically reprojected the step into the nearest safe region. As shown in the accompanying video and Figure \ref{fig:roughTerrainUI}, the robot frequently corrected user steps that were not on planar surfaces, ensuring that each foot landed securely on the cinder blocks.

\section{Conclusions}
This paper introduced a teleoperation framework for humanoid robots that enables real-time, user-driven stepping on both flat and rough terrain. The system retargets user footstep intentions to feasible robot foot placements, continuously refining these estimates as the user moves. It also incorporates a terrain adaptation module that uses onboard depth sensing to construct elevation maps, allowing the robot to autonomously adjust its steps for stability on uneven surfaces.

Experiments demonstrated that the robot can closely mirror the user’s stepping on flat ground and reliably traverse challenging cinder block terrain, correcting for user steps that would otherwise be unstable. These results highlight the potential of anticipatory and adaptive footstep streaming to enhance the responsiveness and robustness of bipedal robot teleoperation.

Future work could further reduce the delay between user and robot stepping by developing controllers that allow the robot to initiate the swing phase concurrently with the transfer phase, potentially leveraging learning-based \cite{he2024} or model predictive control techniques \cite{mastalli2023}. Additionally, exploring the use of robots with lower leg inertia may enable faster stepping and improved synchronization between the user and the robot.

\newpage
\bibliography{main}

\end{document}

%% file: SwingTrajectoryCollisionAvoidance.tex
\subsection{Collision-Free Swing Foot Trajectory}

Accurate footstep placement on uneven terrain requires generating swing foot trajectories that avoid collisions with obstacles. Such collisions may occur when a new footstep is placed at a height different from the previous one, particularly if the swing foot does not sufficiently clear potential obstacles. To ensure both safe clearance and smooth transitions between steps, the swing foot trajectory must be carefully modified.

When a height difference is detected between two consecutive footsteps, the swing foot trajectory is adjusted by introducing two intermediate waypoints, denoted as $P_{wp1}$ and $P_{wp2}$. These waypoints are determined using predefined proportions, $\alpha_{wp1}$ and $\alpha_{wp2}$, which specify their relative positions along the path from the initial foot position ($P_{initial}$) to the goal position ($P_{goal}$), where the new footstep will be placed. The horizontal (XY-plane) positions of the waypoints are calculated through linear interpolation between the initial and goal positions as follows:

\begin{equation}
    \mathbf{P}_{\mathrm{wp(i)}}^{XY} = (1-\alpha_{\mathrm{wp(i)}}) \cdot \mathbf{P}_{\mathrm{initial}}^{XY} + \alpha_{\mathrm{wp(i)}} \cdot \mathbf{P}_{\mathrm{goal}}^{XY},
\end{equation}

where $wp(i)$ refers to each waypoint, and the sum $\alpha_{\mathrm{wp1}} + \alpha_{\mathrm{wp2}} = 1$.

The vertical (Z-axis) component of the swing foot trajectory is adjusted to avoid obstacles, with the adjustment depending on whether the foot is stepping up to a higher surface or down to a lower one. Specifically, the Z-coordinates of the waypoints are computed as:
{\small
\begin{align}
    P_{\mathrm{wp1}}^{Z} &= h_{\mathrm{swing}} + (1-\mathbf{1}_{\mathrm{up}} \cdot \alpha_{\mathrm{wp1}}) \cdot P_{\mathrm{initial}}^Z + \mathbf{1}_{\mathrm{up}} \cdot\alpha_{\mathrm{wp1}} \cdot P_{\mathrm{goal}}^{Z}, \\
    P_{\mathrm{wp2}}^{Z} &= h_{\mathrm{swing}} + (1-\mathbf{1}_{\mathrm{down}} \cdot \alpha_{\mathrm{wp2}}) \cdot P_{\mathrm{initial}}^Z + \mathbf{1}_{\mathrm{down}} \cdot\alpha_{\mathrm{wp2}} \cdot P_{\mathrm{goal}}^{Z},
\end{align}
}
where $\mathbf{1}_{\mathrm{up}}$ equals $1$ when stepping up and $0$ otherwise, while $\mathbf{1}_{\mathrm{down}}$ is defined conversely. The parameter $h_{\mathrm{swing}}$ represents the predefined height that the swing foot should reach during its trajectory.

Once the four key positions—$\mathbf{P}_{\mathrm{initial}}$, $\mathbf{P}_{\mathrm{wp1}}$, $\mathbf{P}_{\mathrm{wp2}}$, and $\mathbf{P}_{\mathrm{goal}}$—are established, a cubic spline is used to generate a smooth swing foot trajectory passing through these points. The timing of each segment is determined based on the total swing duration ($T_{\mathrm{swing}}$) and the waypoint proportions. In this approach, $\alpha_{\mathrm{wp1}}$ is set to $0.15$ and $\alpha_{\mathrm{wp2}}$ to $0.85$, ensuring that the waypoints are appropriately spaced along the trajectory.

This method effectively combines obstacle avoidance with smooth and natural foot motion, which is essential for stable walking over complex and uneven surfaces.